\documentclass[sigconf]{acmart}
\AtBeginDocument{%
  }

\usepackage{bm}
\usepackage{graphicx}
\usepackage{tikz}
\usepackage{color}
\usepackage{multirow}
\usepackage{multicol}
\usepackage{colortbl}
\usepackage{hyperref}

\copyrightyear{2025}
\acmYear{2025}
\setcopyright{acmlicensed}
\acmConference[MM '25] {Proceedings of the 33rd ACM International Conference on Multimedia}{October 27--31, 2025}{Dublin, Ireland.}
\acmBooktitle{Proceedings of the 33rd ACM International Conference on Multimedia (MM '25), October 27--31, 2025, Dublin, Ireland}
\acmDOI{10.1145/3746027.3754882}
\acmISBN{979-8-4007-2035-2/2025/10}

\begin{document}

\title{SPHERE: Semantic-PHysical Engaged REpresentation for 3D Semantic Scene Completion}

\author{Zhiwen Yang}
\orcid{0000-0002-0416-0576}
\affiliation{%
  \institution{Peking University}
  \department{Wangxuan Institute of Computer Technology}
  \city{Beijing}
  \country{China}
}
\email{yangzhiwen@pku.edu.cn}

\author{Yuxin Peng}
\authornote{Corresponding author.}
\orcid{0000-0001-7658-3845}
\affiliation{%
  \institution{Peking University}
  \department{Wangxuan Institute of Computer Technology}
  \city{Beijing}
  \country{China}
}
\email{pengyuxin@pku.edu.cn}

\renewcommand{\shortauthors}{Zhiwen Yang and Yuxin Peng}

\begin{abstract}
Camera-based 3D Semantic Scene Completion (SSC) is a critical task in autonomous driving systems, assessing voxel-level geometry and semantics for holistic scene perception. While existing voxel-based and plane-based SSC methods have achieved considerable progress, they struggle to capture physical regularities for realistic geometric details. On the other hand, neural reconstruction methods like NeRF and 3DGS demonstrate superior physical awareness, but suffer from high computational cost and slow convergence when handling large-scale, complex autonomous driving scenes, leading to inferior semantic accuracy. To address these issues, we propose the \textbf{S}emantic-\textbf{PH}ysical \textbf{E}ngaged \textbf{RE}presentation (\textbf{SPHERE}) for camera-based SSC, which integrates voxel and Gaussian representations for joint exploitation of semantic and physical information. First, the Semantic-guided Gaussian Initialization (SGI) module leverages dual-branch 3D scene representations to locate focal voxels as anchors to guide efficient Gaussian initialization. Then, the Physical-aware Harmonics Enhancement (PHE) module incorporates semantic spherical harmonics to model physical-aware contextual details and promote semantic-geometry consistency through focal distribution alignment, generating SSC results with realistic details. Extensive experiments and analyses on the popular SemanticKITTI and SSCBench-KITTI-360 benchmarks validate the effectiveness of SPHERE. The code is available at \textcolor{blue}{\href{https://github.com/PKU-ICST-MIPL/SPHERE_ACMMM2025}{https://github.com/PKU-ICST-MIPL/SPHERE\_ACMMM2025}}.

\end{abstract}

\ccsdesc[500]{Computing methodologies~Scene understanding}
\ccsdesc[500]{Computing methodologies~Reconstruction}

\keywords{Semantic Scene Completion; Semantic-Geometry Consistency; Semantic Spherical Harmonics}
\begin{teaserfigure}
  \includegraphics[width=\textwidth]{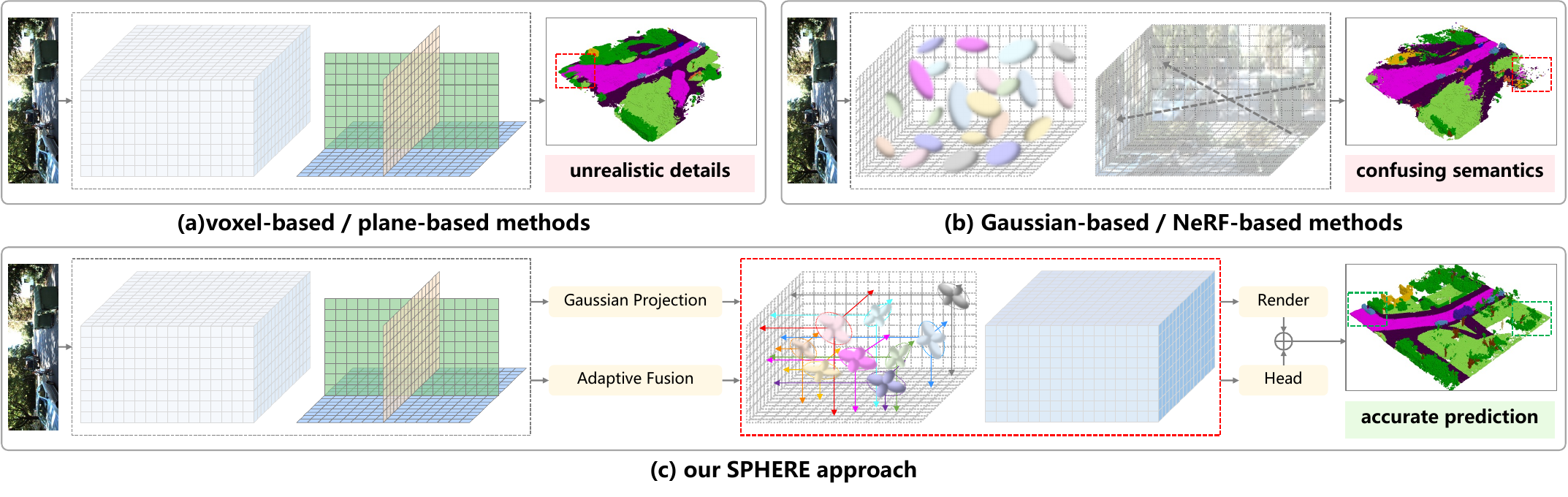}
  \caption{Comparison among different representations for SSC. (a) Voxel-based and plane-based methods demonstrate effectiveness in semantic accuracy, but struggle with modeling real-world physical regularities, leading to unrealistic details. (b) Due to spatial redundancy, Gaussian-based and NeRF-based methods suffer from high computation cost and slow convergence, resulting in confusing semantics. (c) Our SPHERE approach exploits semantic-geometry consistency via efficient integration of voxel and Gaussian representations, generating accurate SSC predictions.}
  \label{fig:motivation}
\end{teaserfigure}


\maketitle

\section{Introduction}
In autonomous driving systems, 3D Semantic Scene Completion (SSC)~~\cite{roldao20223d} is a crucial task for holistic scene perception, inferring voxel-level geometry and semantics for several downstream tasks. To deal with the inherent complexity of large-scale outdoor scenarios, LiDAR-based SSC methods~\cite{roldao2020lmscnet, cheng2021s3cnet, rist2021semantic, yan2021sparse, xia2023scpnet, zhang2023attention, ye2022unsupervised, li2023lode, lee2024semcity} have been the predominant solutions, taking advantage of explicit 3D geometric structural information entailed in the point cloud data. Despite their effectiveness, the inherently high cost and limited scalability of LiDAR sensors constrain practical applications of LiDAR-based methods in real-world deployment. In light of this issue, multi-view images~\cite{miao2023occdepth, yao2023ndc, tong2023scene, zheng2024multi} have garnered increasing attention as a more cost-effective and flexible input modality.

The pioneering camera-based SSC method, MonoScene~\cite{cao2022monoscene}, extracts 2D features from input RGB images and lifts them into dense 3D volume features through depth-based projection. Following this, existing camera-based SSC methods fall mainly into two paradigms: voxel-based and plane-based methods. Voxel-based SSC methods~\cite{zhang2023occformer, li2023voxformer} discretize 3D scenes into voxels and assign extracted image features to each voxel, and progress has been achieved in exploiting scene-from-instance feature interactions~\cite{jiang2024symphonize}, and employing a dense-sparse-dense strategy for hybrid guidance~\cite{mei2024camera}. In addition, plane-based SSC methods adopt Bird's-Eye-View (BEV)~\cite{huang2021bevdet} and Tri-Perspective View (TPV)~\cite{huang2023tri} features for efficient scene representations. Subsequent research further improves the quality of scene representation through camera-aware depth estimation and refinement~\cite{li2023bevdepth}, dynamic temporal stereo information~\cite{li2023bevstereo}, and explicit-implicit compact feature projection~\cite{ma2024cotr}.

While existing SSC methods have demonstrated promising performance in voxel-wise semantic prediction, they frequently fail to capture the underlying physical regularities in the real world, leading to unrealistic geometric details in scene completion results, as illustrated in Figure~\ref{fig:motivation} (a). On the other hand, neural reconstruction methods like NeRF~\cite{mildenhall2021nerf} and 3DGS~\cite{kerbl20233d} excel at learning realistic geometry details, but suffer from high computational expense and slow convergence in large-scale complex autonomous driving scenes, resulting in confusing semantics, as shown in Figure~\ref{fig:motivation} (b). The above challenges naturally raise a critical question: \textbf{\textit{How can we achieve both accurate semantics and realistic geometry, efficiently and effectively?}}

In this paper, we propose the \textit{Semantic-PHysical Engaged REpresentation} (\textbf{SPHERE}) framework for 3D semantic scene completion as an alternative to answer this question, which combines voxel and Gaussian representations to jointly exploit semantic structures and physical regularities for accurate SSC prediction with accurate semantics and realistic geometry, as illustrated in Figure~\ref{fig:motivation} (c).
Our motivation derives from the observation that a large proportion of empty voxels in autonomous driving scenes cause severe spatial redundancy in neural representation methods, directly leading to high computational expense and slow convergence speed.
In light of this, the \textit{Semantic-guided Gaussian Initialization} (\textbf{SGI}) module is proposed to utilize dual-branch 3D scene representations to identify focal voxels with discriminative semantics as anchors, guiding efficient and effective Gaussian initialization.
Then, to fully leverage the powerful physical structure modeling capability of Gaussian representations, we introduce the \textit{Physical-aware Harmonics Enhancement} (\textbf{PHE}) module, which devises semantic spherical harmonics to model physical-aware contextual structures, then enhances semantic-geometry consistency via focal distribution alignment between voxel and Gaussian representations, thereby generating realistic semantic scene completion results.
Extensive experiments and analyses on the SemanticKITTI and SSCBench-KITTI-360 datasets validate the effectiveness of SPHERE, demonstrating improved semantic accuracy and geometric fidelity over state-of-the-art methods.

In summary, the main contributions of our work are as follows:
\begin{itemize}
    \item We propose the SPHERE framework, which efficiently integrates voxel and Gaussian representations to jointly exploit semantic information and physical regularities, generating SSC results with accurate semantics and realistic geometry.
    \item SGI leverages dual-branch 3D scene representations to identify focal voxels with discriminative semantics as anchors, enabling effective and efficient Gaussian initialization to alleviate spatial redundancy.
    \item PHE module incorporates semantic spherical harmonics for improved physical-aware contextual details, and promotes focal distribution alignment between voxel and Gaussian representations to enhance semantic-geometric consistency.
\end{itemize}

\begin{figure*}[t]
\centering
\includegraphics[width=\textwidth]{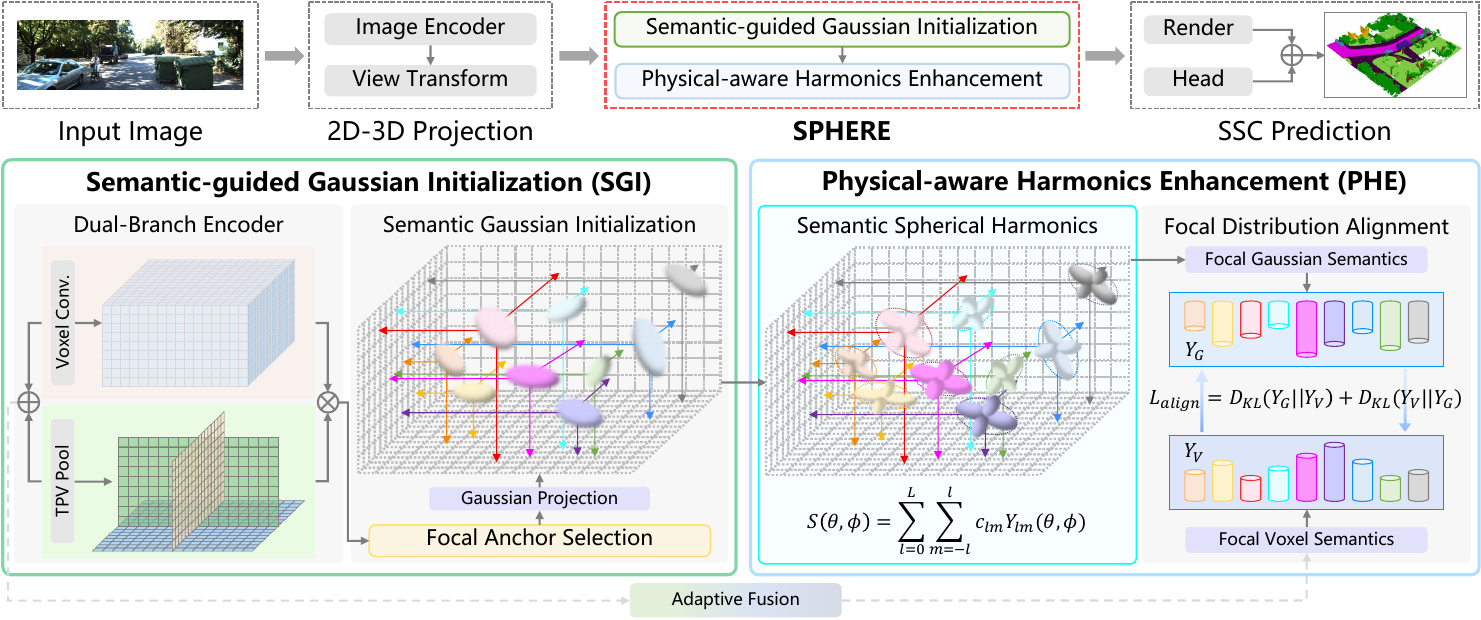}
\caption{The overall architecture of our SPHERE approach. The Semantic-guided Gaussian Initialization (SGI) module leverages a dual-branch encoder to exploit local and global semantics, and then selects focal voxels as anchors for effective and efficient Gaussian initialization. The Physical-aware Harmonics Enhancement (PHE) module incorporates semantic spherical harmonics to model physical-aware contextual details, then enhances semantic-geometry consistency via focal distribution alignment between the voxel and Gaussian semantics.}
\label{fig:framework}
\end{figure*}

\section{Related Work}
In this section, we provide a brief review of pertinent literature related to our work, focusing on camera-based SSC methods and neural representation methods in 3D perception.

\subsection{Camera-based Semantic Scene Completion}
\textbf{Voxel-based SSC} methods discretize 3D space into voxels and assign image features to each voxel for scene perception.  MonoScene~\cite{cao2022monoscene} is a pioneering effort that reconstructs 3D scenes from RGB images by projecting image features to 3D space along optical rays. OccFormer~\cite{zhang2023occformer} introduces a dual-path transformer architecture with a mask-wise prediction paradigm, effectively capturing spatial dependencies for improved completion accuracy. VoxFormer~\cite{li2023voxformer} presents a two-stage framework that integrates class-agnostic query proposals with class-specific semantic segmentation, facilitating the propagation of discriminative semantic information from seed voxels across the entire scene for enhanced completion. Symphonize~\cite{jiang2024symphonize} introduces a scene-from-instance framework that enhances the interplay between image and volumetric representations, enabling more precise scene completion. SGN~\cite{mei2024camera} presents a one-stage SSC framework employing a dense-sparse-dense design, which enhances segmentation boundary clarity and improves the accuracy of 3D semantic scene completion. CGFormer~\cite{yu2024context} introduces a context- and geometry-aware voxel transformer that utilizes context-dependent queries and multiple 3D representations to enhance scene completion performance.
\\
\textbf{Plane-based SSC} methods compress the 3D space along certain dimensions for efficient view transformation and scene representation. A pioneering approach in this domain is the Lift-Splat-Shoot (LSS)~\cite{philion2020lift} framework, which establishes an end-to-end pipeline by first "lifting" images into feature frustums, then "splatting" these frustums onto a rasterized BEV grid, and finally "shooting" template trajectories into a BEV cost map for downstream tasks. Building upon the LSS paradigm, BEVDet~\cite{huang2021bevdet} introduces a holistic BEV-based framework for scene understanding, comprising an Image-view Encoder, a View Transformer, a BEV Encoder, and a Task-specific Head. Subsequent achievements have been made in further enhancing the BEV-based representations, incorporating techniques such as inverse perspective mapping~\cite{hu2021fiery}, angle and radius aware rasterization~\cite{liu2023vision}, and cross-attention layers with BEV queries~\cite{li2022bevformer}. BEVDepth~\cite{li2023bevdepth} enhances depth learning by introducing a camera-aware depth estimation module alongside a depth refinement module, improving the accuracy of depth predictions. Building on this, BEVStereo~\cite{li2023bevstereo} advances depth estimation by incorporating dynamic temporal stereo information. Beyond BEV, TPVFormer~\cite{huang2023tri} introduces a Tri-Perspective View (TPV) representation, which decomposes voxel grids into three orthogonal planes, allowing efficient scene encoding.

Existing camera-based semantic scene completion methods have made great progress in predicting voxel-wise semantics, but fall short of the ability to capture physical regularities for realistic geometry details. In contrast, our SPHERE employs Gaussian representations around focal regions for the exploitation of physical-aware structures, facilitating realistic geometric details in the semantic scene completion results.

\subsection{Neural Representation in 3D Perception}
Neural representation techniques have emerged as powerful tools for modeling geometric details. Neural Radiance Field (NeRF)~\cite{mildenhall2021nerf} represents the 3D scene with implicit continuous functions, mapping point coordinates and view directions to target properties, and employs ray-marching for volume rendering. 3D Gaussian Splatting (3DGS)~\cite{kerbl20233d} explicitly initializes a set of Gaussian distributions to represent the geometry and appearance within the scene, improving efficiency and scalability with a rasterization-based rendering pipeline. Omni-Scene~\cite{wei2024omni} takes advantage of both pixel and volume-based representations and proposes an Omni-Gaussian representation for improved ego-centric reconstruction and novel view synthesis performance. RenderOcc~\cite{pan2024renderocc} extracts a NeRF-style 3D scene representation and utilizes 2D rendering techniques for 3D supervision from 2D semantics and depth labels. GaussianFormer~\cite{huang2024gaussianformer, huang2024probabilistic} proposes image-to-Gaussian mapping with distribution-based initialization and Gaussian-to-voxel splatting with probabilistic Gaussian superposition for 3D semantic scene completion.

Although superior in learning physical structures and geometric details, neural representation methods suffer from high computational cost and slow convergence speed in autonomous driving scenes due to spatial redundancy. In contrast, our SPHERE approach identifies focal voxels as anchors to guide efficient and effective Gaussian initializations, alleviating spatial redundancy while enabling semantic scene completion with accurate semantics and realistic geometry.

\section{Methodology}
\subsection{Overview}
Figure~\ref{fig:framework} demonstrates the overall framework of our Semantic-Physical Engaged REpresentation (SPHERE) approach, consisting of two key components: (1) a Semantic-guided Gaussian Initialization (SGI) module that employs dual-branch 3D representation to select focal voxel anchors for efficient and effective Gaussian initialization; (2) a Physical-aware Harmonics Enhancement (PHE) module that adopts semantic spherical harmonics for improved physical-aware contextual details and aligns focal voxel and Gaussian semantic distribution for enhanced semantic-geometry consistency.
\\
\textbf{Problem Setup.} Given a pair of stereo images $I_{l}^{\rm rgb}, I_{r}^{\rm rgb}$, semantic scene completion (SSC) aims to infer the geometry and semantics of 3D scenes in front. The voxelized output is structured with grid size $Y\in\mathbb{R}^{X\times Y\times Z}$, where $X, Y, Z$ correspond to the grid's length, width, and height, respectively. Each voxel is classified as either empty, denoted by $c_0$ or occupied by one of the semantic classes in $c\in\{c_1,\cdots,c_N\}$, where $N$ represents the number of semantic classes. Essentially, SSC aims to train a model $V=\Theta(I_{l}^{\rm rgb}, I_{r}^{\rm rgb})$ that can generate a 3D semantic prediction $V$ aligning closely with the ground truth $\Bar{V}$.
\\
\textbf{3D Semantic Gaussian Representation.} 3D semantic Gaussian representation initializes a set of $P$ Gaussian distributions $\mathcal{G}=\{G_{i}\}_{i=1}^{P}$, where each $G_i$ covers a local area determined by its mean $m_i$, scale $s_i$, rotation $r_i$, opacity $\alpha_i$ and semantic $c_i$. These Gaussian distributions contribute to voxel semantics in an additive manner:
\begin{equation}
    y(\mathbf{x}; \mathcal{G}) = \sum\limits_{i=1}^{P}g_{i}(\mathbf{x};m_i,s_i,r_i,\alpha_i,c_i)
\end{equation}
where $y(\mathbf{x}; \mathcal{G})$ denotes the predicted occupancy concerning all Gaussian distributions at location $\mathbf{x}$, and $g_{i}(x;\cdot)$ indicates the contribution of the $i$-th semantic Gaussian. The contribution is evaluated based on the Gaussian properties:
\begin{equation}
    g_i(\mathbf{x};G_i) = \alpha_i \cdot \exp(-\frac{1}{2}(x-m_i)^{\rm{T}}\Sigma_{i}^{-1}(x-m_i))c_i
\end{equation}
\begin{equation}
    \Sigma_i=RSS^{\rm{T}}R^{\rm{T}},\quad S=\rm{diag}(s),\quad R=\rm{q2r(r)}
\end{equation}
where $\Sigma_i$ is the covariance matrix, $S_i$ represents the diagonal scale matrix, and $R_i$ denotes the rotation matrix.

\subsection{Semantic-guided Gaussian Initialization}
\textbf{Design rationale.} The large proportion of empty voxels in autonomous driving scenes causes severe spatial redundancy in neural representation techniques like NeRF~\cite{mildenhall2021nerf} and 3DGS~\cite{kerbl20233d} for modeling non-occupied areas, resulting in high computational expense and slow convergence speed. In light of this, the SGI module first leverages dual-branch 3D representations for global and local semantic information, and then identifies focal voxels as anchors for efficient and effective Gaussian initialization.
\\
\textbf{Dual-Branch Encoder.} Notice that TPV~\cite{huang2023tri} representations compress the full voxel representations along three axes respectively, thereby focusing more on global semantics, while dense voxel representations highlight local semantic details. We leverage both voxel and TPV representations for comprehensive aggregation of local and global semantic information. Given the initial 3D features $\textbf{F}_{\rm 3D}$ projected from image features $\textbf{F}_{\rm img}$, the voxel representation is derived through multiple 3D convolution layers followed by a 3D pyramid network:
\begin{equation}
    \textbf{F}_{\rm voxel} = {\rm FPN}_{\rm 3D}\left({\rm Conv_{\rm 3D}\left(\textbf{F}_{\rm 3D}\right)}\right)
\end{equation}
where $\textbf{F}_{\rm voxel}\in\mathbb{R}^{C\times X\times Y\times Z}$ denotes the voxel representation with local semantic details. On the other hand, the TPV representations are obtained with three spatial pooling layers to compress 3D features along the perpendicular axes. The compressed feature maps are further passed through 2D Convolution layers to refine global semantics:
\begin{equation}
    \textbf{F}_{p} = {\rm Conv_{\rm 2D}}\left(\rm{Pool}_{p}\left(\textbf{F}_{\rm 3D}\right)\right),\quad p\in\{XY,YZ,ZX\}
\end{equation}
where $\textbf{F}_{XY}\in\mathbb{R}^{C\times X\times Y},\textbf{F}_{YZ}\in\mathbb{R}^{C\times Y\times Z},\textbf{F}_{ZX}\in\mathbb{R}^{C\times X\times Z}$ indicate the compressed features and the overall TPV representations are viewed as a list $\mathcal{F}_{\rm TPV}=[\textbf{F}_{XY}, \textbf{F}_{YZ}, \textbf{F}_{ZX}]$.
\\
\begin{figure}[t]
\centering
\includegraphics[width=1.0\columnwidth]{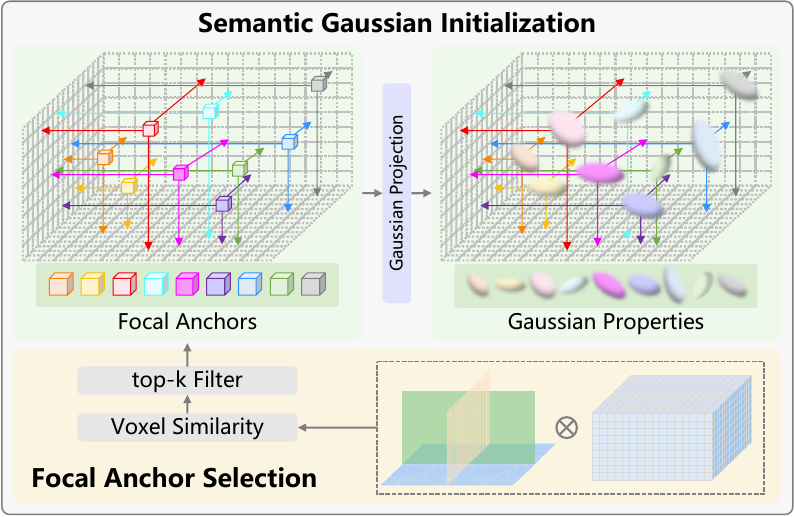}
\caption{Illustration of the Semantic Gaussian Initialization process. The voxel-wise feature similarities are first computed between the voxel and TPV features to select top-$k$ focal voxels as anchors. Then, a Gaussian projection layer is employed to generate Gaussian properties according to the semantics and positions of focal anchors.}
\label{fig:sgi}
\end{figure}
\textbf{Focal Anchor Selection.} To address spatial redundancy, we identify focal voxels with discriminative semantics as anchors for efficient and effective Gaussian initialization around occupied areas. Specifically, we first calculate the similarity score across local semantics embedded in $\textbf{F}_{\rm voxel}$ and global semantics embedded in $\mathcal{F}_{\rm TPV}$:
\begin{equation}
    M_{\rm sim} = {\rm sim}\left(\textbf{F}_{\rm voxel}, \mathcal{W}\odot\mathcal{F}_{\rm TPV}\right)
\end{equation}
where $M_{\rm sim}\in\mathbb{R}^{X\times Y\times Z}$ represents the semantic similarity map, ${\rm sim}(\cdot,\cdot)$ denotes the voxel-wise cosine similarity function, and $\mathcal{W}$ is the aggregation weight list for TPV representations. Then, we select top-$k$ focal voxels with the highest similarity scores as anchors for Gaussian initializations:
\begin{equation}
\begin{aligned}
    \textbf{P}_{\rm anc} &= {\rm top}_{k}(M_{\rm sim}, K) \\ 
    \textbf{F}_{\rm anc} &= (\textbf{F}_{\rm voxel}+\mathcal{W}\odot\mathcal{F}_{\rm TPV})[\textbf{P}_{\rm anc}]
\end{aligned}
\end{equation}
where $\textbf{P}_{\rm anc}\in\mathbb{R}^{K\times 3}, \textbf{F}_{\rm anc}\in\mathbb{R}^{K\times C}$ represent the positions and semantic features of the selected $K$ focal voxels respectively.
\\
\textbf{Semantic Gaussian Initialization.} The focal voxels with high similarity scores possess consistent local and global semantics, thereby serving as ideal anchors for Gaussian initializations. As illustrated in Figure~\ref{fig:sgi}, the initialization process is formulated as follows:
\begin{equation}
    [\textbf{o},\textbf{s},\textbf{r},\bm{\alpha}] = {\rm MLP_{GS}}(\textbf{F}_{\rm anc})
\end{equation}
where offset $\textbf{o}$, scale $\textbf{s}$, rotation $\textbf{r}$, and opacity $\bm{\alpha}$ are Gaussian properties projected from discriminative semantics of focal anchors. Then, the corresponding activation functions are applied for normalization:
\begin{equation}
\begin{aligned}
    \textbf{m} &= \textbf{P}_{\rm anc} + {\rm tanh}(\textbf{o}) \\
    [\textbf{s}, \bm{\alpha}] &= {\rm sigmoid}([\textbf{s}, \bm{\alpha}]) \\
    \textbf{r} &= {\rm norm}\left(\textbf{r}\right)\\
\end{aligned}
\end{equation}
By concatenating semantic features and normalized properties, the initialized Gaussians $\mathcal{G}=[\textbf{F}_{\rm anc}, \textbf{m},\textbf{s},\textbf{r},\bm{\alpha}]\in\mathbb{R}^{K\times (C+11)}$ efficiently cover focal regions with discriminative semantics.

\subsection{Physical-aware Harmonics Enhancement}
\textbf{Design rationale.} To generate realistic geometry structures without harming semantic accuracy, it is crucial to model physical-ware geometric details in accordance with voxel semantics. Therefore, the PHE module employs semantic spherical harmonics for modeling contextual details, and ensures semantic-geometry consistency with focal distribution alignment.
\begin{figure}[t]
\centering
\includegraphics[width=1.0\columnwidth]{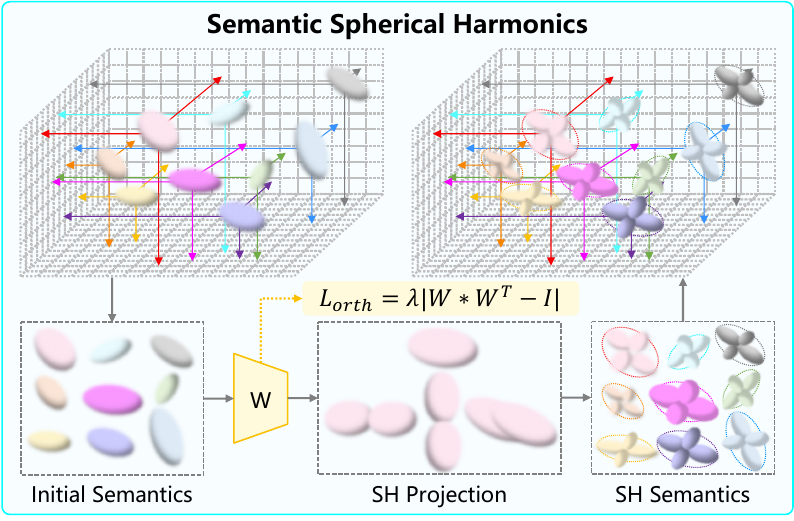}
\caption{Illustration of the Semantic Spherical Harmonics. The initial Gaussian semantics are passed through SH projection to cope with SH coefficients, exploiting physical-aware contextual details. Furthermore, we employ an orthogonal loss on the projection matrix, considering that spherical harmonics form a complete set of orthogonal functions.}
\label{fig:ssh}
\end{figure}

\textbf{Semantic Spherical Harmonics.} Spherical Harmonics (SH) demonstrates superior performance in RGB-based color rendering~\cite{kerbl20233d, wu20244d}, showcasing its powerful capability to model spatial variations and complex geometry. To leverage this capability for SSC, we employ the Semantic Spherical Harmonics (SSH), as shown in Figure~\ref{fig:ssh}, which incorporates semantic priors into SH basis functions:
\begin{equation}
    S(\theta, \phi) = \sum\limits_{l=0}^{L}\sum\limits_{m=-l}^{l}c_{lm}Y_{lm}(\theta, \phi)
\end{equation}
where $l, m$ are the degree and order controlling the level of details, $(\theta, \phi)$ indicates the spherical coordinates, $Y_{lm}(\theta, \phi)$ denotes the SH basis function, and $c_{lm}$ represents the expanded semantics. To cope with the Semantic Spherical Harmonics of $L$ degrees, we employ 1D convolution layers to expand focal voxel semantics:
\begin{equation}
    \{c_{lm}\} = \textbf{F}_{\rm exp} = Conv_{\rm 1D}(\textbf{F}_{\rm anc})
    \label{exp}
\end{equation}
where $\textbf{F}_{\rm exp}\in\mathbb{R}^{K\times 2^{L}C_{N}}$ corresponds to $2^L$ expanded semantics $c_{lm}$ in the Semantic Spherical Harmonics. Considering that spherical harmonics form a complete set of orthogonal functions, to further enhance the capability of Semantic Spherical Harmonics in modeling contextual details, different degrees and orders $(l,m)$ should correspond to distinct semantics. Therefore, we employ the soft orthogonality loss for supervision:
\begin{equation}
    \mathcal{L}_{\rm orth} = \lambda\left|W*W^{\rm T} - I\right|
\end{equation}
where $W$ denotes the weight matrix in equation~(\ref{exp}), $I$ is the identity matrix, and $\lambda$ is the regularization parameter. By promoting the orthogonality across projection weights, the expanded semantics $\{c_{lm}\}$ learn to capture distinct semantic information, thereby better modeling semantic-aware spatial variations and geometric details. 
\\
\textbf{Focal Distribution Alignment.} With the dual-branch 3D representations and sematic-guided Gaussian representations, we generate semantic scene completion predictions with task head and Gaussian superposition respectively:
\begin{equation}
\begin{aligned}
    \textbf{V}_{\rm voxel} &= {\rm Head}\left(\textbf{F}_{\rm voxel}+\mathcal{W}\odot\mathcal{F}_{\rm TPV}\right) \\
    \textbf{V}_{\rm gauss}[\textbf{x}] &= y(\textbf{x};[S_{\theta,\phi}(\textbf{F}_{\rm exp}),\textbf{m},\textbf{s},\textbf{r},\bm{\alpha}])
\end{aligned}
\end{equation}
where $\textbf{V}_{\rm voxel}$ focuses on semantic accuracy, and $\textbf{V}_{\rm gauss}$ emphasizes geometry details. To promote the consistency between contextual semantics and geometry, we adopt an alignment loss across the predictions concerning focal voxels:
\begin{equation}
\begin{aligned}
    \mathcal{L}_{\rm align} &= D_{\rm KL}\left(\textbf{V}_{\rm voxel}[\textbf{P}_{\rm anc}]||\textbf{V}_{\rm gauss}[\textbf{P}_{\rm anc}]\right) \\
    & + D_{\rm KL}\left(\textbf{V}_{\rm gauss}[\textbf{P}_{\rm anc}]||\textbf{V}_{\rm voxel}[\textbf{P}_{\rm anc}]\right)
\end{aligned}
\end{equation}
where $D_{\rm KL}(\cdot||\cdot)$ is the Kullback-Leibler divergence measuring the difference between two distributions. The aligned predictions are then aggregated to generate the final prediction:
\begin{equation}
    \textbf{V}_{\rm SSC} = \textbf{V}_{\rm voxel} + \textbf{V}_{\rm gauss}
\end{equation}

\subsection{Training Objective}
Following~\cite{cao2022monoscene, mei2024camera}, we adopt the cross-entropy loss $\mathcal{L}_{\rm ce}$, lovasz loss $\mathcal{L}_{\rm lovasz}$~\cite{berman2018lovasz}, and scene-class affinity loss $\mathcal{L}_{\rm scal}$ to supervise the final predictions. The overall loss objective is formulated as follows:
\begin{equation}
    \mathcal{L} = \mathcal{L}_{\rm ce} + \mathcal{L}_{\rm lovasz} + \mathcal{L}_{\rm scal} + \mathcal{L}_{\rm orth} + \mathcal{L}_{\rm align}
\end{equation}

\begin{table*}
  \centering
  \caption{Camera-based 3D semantic scene completion results on the SemanticKITTI~\cite{behley2019semantickitti} validation set. $^{\dagger}$ denotes the methods employing EfficientNetB7~\cite{tan2019efficientnet} as image backbone, and $^{*}$ represents the methods utilizing ResNet50~\cite{he2016deep} as image backbone. Best results are highlighted in bold, and second-best scores are \underline{underlined}.}
  \resizebox{1.0\linewidth}{!}{
  \begin{tabular}{l|cc|ccccccccccccccccccc}
    \toprule
    \multirow{2}{*}{Method} & SC & SSC
    & \rotatebox{90}{\multirow{2}{*}{\textcolor[RGB]{91,155,213}{$\blacksquare$} \textbf{car}{\footnotesize (3.92\%)}}} 
    & \rotatebox{90}{\multirow{2}{*}{\textcolor[RGB]{100,230,245}{$\blacksquare$} \textbf{bicycle}{\footnotesize (0.03\%)}}} 
    & \rotatebox{90}{\multirow{2}{*}{\textcolor[RGB]{30,60,150}{$\blacksquare$} \textbf{motorcycle}{\footnotesize (0.03\%)}}} 
    & \rotatebox{90}{\multirow{2}{*}{\textcolor[RGB]{80,30,180}{$\blacksquare$} \textbf{truck}{\footnotesize (0.16\%)}}}
    & \rotatebox{90}{\multirow{2}{*}{\textcolor[RGB]{0,0,255}{$\blacksquare$} \textbf{other-veh.}{\footnotesize (0.20\%)}}}
    & \rotatebox{90}{\multirow{2}{*}{\textcolor[RGB]{255,30,30}{$\blacksquare$} \textbf{person}{\footnotesize (0.07\%)}}}
    & \rotatebox{90}{\multirow{2}{*}{\textcolor[RGB]{255,37,199}{$\blacksquare$} \textbf{bicyclist}{\footnotesize (0.07\%)}}}
    & \rotatebox{90}{\multirow{2}{*}{\textcolor[RGB]{150,30,90}{$\blacksquare$} \textbf{motorcyclist}{\footnotesize (0.05\%)}}}
    & \rotatebox{90}{\multirow{2}{*}{\textcolor[RGB]{255,0,255}{$\blacksquare$} \textbf{road}{\footnotesize (15.30\%)}}}
    & \rotatebox{90}{\multirow{2}{*}{\textcolor[RGB]{255,150,255}{$\blacksquare$} \textbf{parking}{\footnotesize (1.12\%)}}}
    & \rotatebox{90}{\multirow{2}{*}{\textcolor[RGB]{75,0,75}{$\blacksquare$} \textbf{sidewalk}{\footnotesize (11.13\%)}}}
    & \rotatebox{90}{\multirow{2}{*}{\textcolor[RGB]{175,0,75}{$\blacksquare$} \textbf{other-grnd.}{\footnotesize (0.56\%)}}}
    & \rotatebox{90}{\multirow{2}{*}{\textcolor[RGB]{255,200,0}{$\blacksquare$} \textbf{building}{\footnotesize (14.10\%)}}}
    & \rotatebox{90}{\multirow{2}{*}{\textcolor[RGB]{255,120,50}{$\blacksquare$} \textbf{fence}{\footnotesize (3.90\%)}}}
    & \rotatebox{90}{\multirow{2}{*}{\textcolor[RGB]{0,175,0}{$\blacksquare$} \textbf{vegetation}{\footnotesize (39.3\%)}}}
    & \rotatebox{90}{\multirow{2}{*}{\textcolor[RGB]{135,60,0}{$\blacksquare$} \textbf{trunk}{\footnotesize (0.51\%)}}}
    & \rotatebox{90}{\multirow{2}{*}{\textcolor[RGB]{150,240,80}{$\blacksquare$} \textbf{terrain}{\footnotesize (9.17\%)}}}
    & \rotatebox{90}{\multirow{2}{*}{\textcolor[RGB]{255,240,150}{$\blacksquare$} \textbf{pole}{\footnotesize (0.29\%)}}}
    & \rotatebox{90}{\multirow{2}{*}{\textcolor[RGB]{255,0,0}{$\blacksquare$} \textbf{traf.-sign}{\footnotesize (0.08\%)}}}\\
     & IoU & mIoU & & & & & & & & & & & & & & & & & & &\\
     \midrule
     \multicolumn{22}{l}{\textit{Mono-Input Methods}} \\
    \midrule
    MonoScene$^{\dagger}$~\cite{cao2022monoscene} & 36.86 & 11.08 & 23.26 & 0.61 & 0.45 & 6.98 & 1.48 & 1.86 & 1.20 & 0.00 & 56.52 & 14.27 & 26.72 & 0.46 & 14.09 & 5.84 & 17.89 & 2.81 & 29.64 & 4.14 & 2.25 \\
    TPVFormer$^{\dagger}$~\cite{huang2023tri} & 35.61 & 11.36 & 23.81 & 0.36 & 0.05 & 8.08 & 4.35 & 0.51 & 0.89 & 0.00 & 56.50 & 20.60 & 25.87 & 0.85 & 13.88 & 5.94 & 16.92 & 2.26 & 30.38 & 3.14 & 1.52 \\
    OccFormer$^{\dagger}$~\cite{zhang2023occformer} & 36.50 & 13.46 & 25.09 & 0.81 & 1.19 & 25.53 & 8.52 & 2.78 & 2.82 & 0.00 & 58.85 & 19.61 & 26.88 & 0.31 & 14.40 & 5.61 &  19.63 & 3.93 & 32.62 & 4.26 & 2.86\\
    IAMSSC$^{*}$~\cite{xiao2024instance} & 44.29 & 12.45 & 26.26 & 0.60 & 0.15 & 8.74 & 5.06 & 1.32 & 3.46 & 0.01 & 54.55 & 16.02 & 25.85 & 0.70 & 17.38 & 6.86 & 24.63 & 4.95 & 30.13 & 6.35 & 3.56\\
    \midrule
    \multicolumn{22}{l}{\textit{Temporal-Input Methods}} \\
    \midrule
    VoxFormer$^{*}$~\cite{li2023voxformer} & 44.15 & 13.35 & 26.54 & 1.28 & 0.56 & 7.26 & 7.81 & 1.93 & 1.97 & 0.00 & 53.57 & 19.69 & 26.52 & 0.42 & 19.54 & 7.31 & 26.10 & 6.10 & 33.06 & 9.15 & 4.94\\
    DepthSSC$^{*}$~\cite{yao2023depthssc} &45.84 & 13.28 & 25.94 & 0.35 & 1.16 & 6.02 & 7.50 & \underline{2.58} & \textbf{6.32} & 0.00 & 55.38 & 18.76 & 27.04 & \underline{0.92} & 19.23 & 8.46 & 26.37 & 4.52 & 30.19 & 7.42 & 4.09\\
    Symphonies$^{*}$~\cite{jiang2024symphonize} & 41.92 & 14.89 & 28.68 & \underline{2.54} & \textbf{2.82} & \textbf{20.44} & \underline{13.89} & \textbf{3.52} & 2.24 & 0.00 & 56.37 & 15.28 & 27.58 & 0.95 & 21.64 & 8.40 & 25.72 & 6.60 & 30.87 & 9.57 & 5.76\\
    HASSC$^{*}$~\cite{wang2024not} & 44.58 & 14.74 & 27.33 & 1.07 & 1.14 & 17.06 & 8.83 & 2.25 & \underline{4.09} & 0.00 & 57.23 & 19.89 & 29.08 & \textbf{1.26} & 20.19 & 7.95 & 27.01 & 7.71 & 33.95 & 9.20 & 4.81\\
    H2GFormer$^{*}$~\cite{wang2024h2gformer} & 44.69 & 14.29 & 28.21 & 0.95 & 0.91 & 6.80 & 9.32 & 1.15 & 0.10 & 0.00 & 57.00 & 21.74 & 29.37 & 0.34 & 20.51 & 7.98 & 27.44 & 7.80 & 36.26 & 9.88 & 5.81\\
    CGFormer$^{\dagger}$~\cite{yu2024context} & 45.99 & \underline{16.87} & \underline{34.32} & \textbf{4.61} & \underline{2.71} & \underline{19.44} & 7.67 & 2.38 & 4.08 & 0.00 & \textbf{65.51} & \underline{20.82} & \underline{32.31} & 0.16 & 23.52 & 9.20 & 26.93 & 8.83 & \textbf{39.54} & 10.67 & \underline{7.84}\\
    SGN$^{*}$~\cite{mei2024camera} & \underline{46.21} & 15.32 & 33.31 & 0.61 & 0.46 & 6.03 & 9.84 & 0.47 & 0.10 & 0.00 & 59.10 & 19.05 & 29.41 & 0.33 & \underline{25.17} & \textbf{9.96} & \underline{28.93} & \textbf{9.58} & 38.12 & \textbf{13.25} & 7.32\\
    \midrule
    \rowcolor{gray!10} 
    \textbf{SPHERE}$^{*}$~\textbf{(Ours)} & \textbf{47.91} & \textbf{17.01} & \textbf{34.43} & 0.57 & 0.84 & 14.66 & \textbf{16.62} & 1.02 & 0.85 & 0.00 & \underline{60.32} & \textbf{23.88} & \textbf{32.79} & 0.11 & \textbf{27.60} & \underline{9.71} & \textbf{30.77} & \underline{9.43} & \underline{38.23} & \underline{13.18} & \textbf{8.13} \\
    \bottomrule
  \end{tabular}}
  \label{kitti}
\end{table*}

\section{Experiments}
\subsection{Dataset and Evaluation Metrics}
\textbf{SemanticKITTI Benchmark}~\cite{behley2019semantickitti} is constructed based on the KITTI Odometry Benchmark~\cite{geiger2012we}, comprising 22 sequences of real-world autonomous driving scenarios with 20 semantic categories. The semantic occupancy ground truth is generated as $256\times 256\times 32$ voxel grids with a resolution of 0.2$m$, within the spatial range of $[0\sim 51.2m, -25.6\sim25.6m, -2\sim4.4m]$. Following official protocols,  sequences 00–07 and 09–10 are designated for training, sequence 08 for validation, and sequences 11–21 for testing.
\\
\textbf{SSCBench-KITTI-360 Benchmark}~\cite{li2023sscbench} consists of 9 densely annotated sequences in urban driving environments, which are split into a training set with 8,487 frames from sequences (00, 02-05, 07), a validation set with 1,812 frames from sequence (06), and a test set with 2,566 frames from sequence (09). Semantic annotations across 19 classes cover a spatial region of $[0m, 51.2m]$ in the forward direction, $[-25.6m, 25.6m]$ laterally, and $[-2.0m, 4.4m]$ vertically, which is uniformly discretized into $256 \times 256 \times 32$ voxel grids with a resolution of 0.2$m$.
\\
\textbf{Evaluation Metrics.}
Following~\cite{yu2024context, mei2024camera}, we adopt the Intersection over Union (IoU) of occupied voxels as the evaluation metric for the class-agnostic scene completion (SC). We also report the mean Intersection over Union (mIoU) with respect to all semantic classes to measure the performance of the semantic scene completion (SSC).
\begin{equation}
\begin{aligned}
    {\rm IoU} &= \dfrac{TP}{TP+FP+FN}\\
    {\rm mIoU} &= \dfrac{1}{C}\sum\limits_{c=1}^{C}\dfrac{TP_c}{TP_{c}+FP_{c}+FN_{c}}
\end{aligned}
\end{equation}
where $TP, FP, FN$ represent the number of true positive, false positive, and false negative occupancy predictions, and $C$ stands for the total number of classes.

\subsection{Implementation Details}
\textbf{Network Architecture.}
Following previous methods~\cite{mei2024camera, li2023voxformer, wang2024not}, we adopt the ResNet-50~\cite{he2016deep} network with FPN~\cite{lin2017feature} as the image encoder, generating 2D feature maps with 1/16 of the input resolution. The feature dimension of both 2D and 3D representations $C$ is set to 128. The grid size of 3D feature volume is $(X,Y,Z)=(128,128,16)$, and the final SSC predictions are upsampled to the resolution of $256\times 256\times32$.
In SGI, we employ the dot production to compute voxel-wise feature similarity and select $K=1024$ focal anchors for Gaussian initialization. In the PHE module, we set the regularization parameter for the soft orthogonality loss as $\lambda=1e-6$.
\\
\textbf{Training Setup.}
The RGB images from cam2 are cropped into size of $1220\times370$ as input for the SemanticKITTI dataset, and the RGB images from cam1 are resized to $1408\times376$ for the SSCBench-KITTI-360 dataset. We train SPHERE for 24 epochs on 4 NVIDIA A6000 GPUs, with a total batch size of 4. About 20 GB of GPU memory is consumed on each GPU while training. We employ a self-distillation training strategy~\cite{wang2024not} for more robust and efficient training process. The AdamW~\cite{loshchilov2017decoupled} optimizer is adopted with an initial learning rate of 2e-4 and a weight decay of 1e-2.
\vspace{-1ex}

\begin{table*}
  \centering
  \caption{Camera-based 3D semantic scene completion results on the SSCBench-KITTI-360~\cite{behley2019semantickitti} test set. $^{\dagger}$ denotes the methods employing EfficientNetB7~\cite{tan2019efficientnet} as image backbone, and $^{*}$ represents the methods utilizing ResNet50~\cite{he2016deep} as image backbone. Best results are highlighted in bold, and second-best scores are \underline{underlined}.}
  \resizebox{1.0\linewidth}{!}{
  \begin{tabular}{l|cc|cccccccccccccccccc}
    \toprule
    \multirow{2}{*}{Method} & SC & SSC
    & \rotatebox{90}{\multirow{2}{*}{\textcolor[RGB]{91,155,213}{$\blacksquare$} \textbf{car}{\footnotesize (2.85\%)}}} 
    & \rotatebox{90}{\multirow{2}{*}{\textcolor[RGB]{100,230,245}{$\blacksquare$} \textbf{bicycle}{\footnotesize (0.01\%)}}} 
    & \rotatebox{90}{\multirow{2}{*}{\textcolor[RGB]{30,60,150}{$\blacksquare$} \textbf{motorcycle}{\footnotesize (0.01\%)}}} 
    & \rotatebox{90}{\multirow{2}{*}{\textcolor[RGB]{80,30,180}{$\blacksquare$} \textbf{truck}{\footnotesize (0.16\%)}}}
    & \rotatebox{90}{\multirow{2}{*}{\textcolor[RGB]{0,0,255}{$\blacksquare$} \textbf{other-veh.}{\footnotesize (5.75\%)}}}
    & \rotatebox{90}{\multirow{2}{*}{\textcolor[RGB]{255,30,30}{$\blacksquare$} \textbf{person}{\footnotesize (0.02\%)}}}
    & \rotatebox{90}{\multirow{2}{*}{\textcolor[RGB]{255,0,255}{$\blacksquare$} \textbf{road}{\footnotesize (14.98\%)}}}
    & \rotatebox{90}{\multirow{2}{*}{\textcolor[RGB]{255,150,255}{$\blacksquare$} \textbf{parking}{\footnotesize (2.31\%)}}}
    & \rotatebox{90}{\multirow{2}{*}{\textcolor[RGB]{75,0,75}{$\blacksquare$} \textbf{sidewalk}{\footnotesize (6.43\%)}}}
    & \rotatebox{90}{\multirow{2}{*}{\textcolor[RGB]{175,0,75}{$\blacksquare$} \textbf{other-grnd.}{\footnotesize (2.05\%)}}}
    & \rotatebox{90}{\multirow{2}{*}{\textcolor[RGB]{255,200,0}{$\blacksquare$} \textbf{building}{\footnotesize (15.67\%)}}}
    & \rotatebox{90}{\multirow{2}{*}{\textcolor[RGB]{255,120,50}{$\blacksquare$} \textbf{fence}{\footnotesize (0.96\%)}}}
    & \rotatebox{90}{\multirow{2}{*}{\textcolor[RGB]{0,175,0}{$\blacksquare$} \textbf{vegetation}{\footnotesize (41.99\%)}}}
    & \rotatebox{90}{\multirow{2}{*}{\textcolor[RGB]{150,240,80}{$\blacksquare$} \textbf{terrain}{\footnotesize (7.10\%)}}}
    & \rotatebox{90}{\multirow{2}{*}{\textcolor[RGB]{255,240,150}{$\blacksquare$} \textbf{pole}{\footnotesize (0.22\%)}}}
    & \rotatebox{90}{\multirow{2}{*}{\textcolor[RGB]{255,0,0}{$\blacksquare$} \textbf{traf.-sign}{\footnotesize (0.06\%)}}}
    & \rotatebox{90}{\multirow{2}{*}{\textcolor[RGB]{250,150,0}{$\blacksquare$} \textbf{other-struct.}{\footnotesize (4.33\%)}}}
    & \rotatebox{90}{\multirow{2}{*}{\textcolor[RGB]{50,255,255}{$\blacksquare$} \textbf{other-obj.}{\footnotesize (0.28\%)}}}\\
     & IoU & mIoU & & & & & & & & & & & & & & & & & &\\
     \midrule
     \multicolumn{21}{l}{\textit{Mono-Input Methods}} \\
    \midrule
    MonoScene$^{\dagger}$~\cite{cao2022monoscene} & 37.87 & 12.31 & 19.34 & 0.43 & 0.58 & 8.02 & 2.03 & 0.86 & 48.35 & 11.38 & 28.13 & 3.32 & 32.89 & 3.53 & 26.15 & 16.75 & 6.92 & 5.67 & 4.20 & 3.09\\
    GaussianFormer$^{*}$~\cite{huang2024gaussianformer} & 35.38 & 12.92 & 18.93 & 1.02 & 4.62 & 18.07 & 7.59 & 3.35 & 45.47 & 10.89 & 25.03 & 5.32 & 28.44 & 5.68 & 29.54 & 8.62 & 2.99 & 2.32 & 9.51 & 5.14\\
    GaussianFormer2$^{*}$~\cite{huang2024probabilistic} & 38.31 & 13.90 & 21.08 & 2.55 & 4.21 & 12.41 & 5.73 & 1.59 & 54.12 & 11.04 & 32.31 & 3.34 & 32.01 & 4.98 & 28.94 & 17.33 & 3.57 & 5.48 & 5.88 & 3.54\\
    TPVFormer$^{\dagger}$~\cite{huang2023tri} & 40.22 & 13.64 & 21.56 & 1.09 & 1.37 & 8.06 & 2.57 & 2.38 & 52.99 & 11.99 & 31.07 & 3.78 & 34.83 & 4.80 & 30.08 & 17.52 & 7.46 & 5.86 & 5.48 & 2.70\\
    OccFormer$^{\dagger}$~\cite{zhang2023occformer} & 40.27 & 13.81 & 22.58 & 0.66 & 0.26 & 9.89 & 3.82 & 2.77 & 54.30 & 13.44 & 31.53 & 3.55 & 36.42 & 4.80 & 31.00 & 19.51 & 7.77 & 8.51 & 6.95 & 4.60\\
    IAMSSC$^{*}$~\cite{xiao2024instance} & 41.80 & 12.97 & 18.53 & 2.45 & 1.76 & 5.12 & 3.92 & 3.09 & 47.55 & 10.56 & 28.35 & 4.12 & 31.53 & 6.28 & 29.17 & 15.24 & 8.29 & 7.01 & 6.35 & 4.19\\
    \midrule
    \multicolumn{21}{l}{\textit{Temporal-Input Methods}} \\
    \midrule
    VoxFormer$^{*}$~\cite{li2023voxformer} & 38.76 & 11.91 & 17.84 & 1.16 & 0.89 & 4.56 & 2.06 & 1.63 & 47.01 & 9.67 & 27.21 & 2.89 & 31.18 & 4.97 & 28.99 & 14.69 & 6.51 & 6.92 & 3.79 & 2.43\\
    DepthSSC$^{*}$~\cite{yao2023depthssc} & 40.85 & 14.28 & 21.90 & 2.36 & \underline{4.30} & 11.51 & 4.56 & 2.92 & 50.88 & 12.89 & 30.27 & 2.49 & 37.33 & 5.22 & 29.61 & 21.59 & 5.97 & 7.77 & 5.24 & 3.51\\
    Symphonies$^{*}$~\cite{jiang2024symphonize} & 44.12 & 18.58 & \underline{30.02} & 1.85 & \textbf{5.90} & \underline{25.07} & \underline{12.06} & \textbf{8.20} & 54.94 & 13.83 & 32.76 & \underline{6.93} & 35.11 & \underline{8.58} & 38.33 & 11.52 & 14.01 & 9.57 & \underline{14.44} & \underline{11.28}\\
    CGFormer$^{\dagger}$~\cite{yu2024context} & \underline{48.07} & \underline{20.05} & 29.85 & \underline{3.42} & 3.96 & 17.59 & 6.79 & \underline{6.63} & \textbf{63.85} & \underline{17.15} & \textbf{40.72} & 5.53 & \textbf{42.73} & 8.22 & \underline{38.80} & \textbf{24.94} & 16.24 & \textbf{17.45} & 10.18 & 6.77\\
    SGN$^{*}$~\cite{mei2024camera} & 47.06 & 18.25 & 29.03 & \textbf{3.43} & 2.90 & 10.89 & 5.20 & 2.99 & 58.14 & 15.04 & 36.40 & 4.43 & \underline{42.02} & 7.72 & 38.17 & \underline{23.22} & \underline{16.73} & \underline{16.38} & 9.93 & 5.86\\
    \midrule
    \rowcolor{gray!10} 
    \textbf{SPHERE}$^{*}$~\textbf{(Ours)} & \textbf{48.59} & \textbf{20.56} & \textbf{33.08} & 0.32 & 2.14 & \textbf{25.62} & \textbf{12.79} & 5.16 & \underline{60.05} & \textbf{17.21} & \underline{38.15} & \textbf{9.23} & 40.75 & \textbf{9.24} & \textbf{41.97} & 14.51 & \textbf{18.23} & 11.67 & \textbf{17.22} & \textbf{12.84}\\
    \bottomrule
  \end{tabular}}
  \label{kitti360}
\end{table*}

\begin{table}[t]
    \centering
    \caption{Ablation study on the SemanticKITTI validation set for different architectural components of SPHERE.}
    \begin{tabular}{cc|c|c}
    \toprule
    \textbf{SGI} & \textbf{PHE} & \textbf{IoU} & \textbf{mIoU} \\
    \midrule
    \midrule
     & & 44.15 & 13.35 \\
    $\checkmark$ & & 46.23 & 15.37 \\
    $\checkmark$ & $\checkmark$ & \textbf{47.91} & \textbf{17.01} \\
    \bottomrule
    \end{tabular}
    \label{tab:ablation}
\end{table}
\begin{table}[t]
    \centering
    \caption{Ablation study on the SemanticKITTI validation set validating the effectiveness of PHE.}
    \begin{tabular}{cc|c|c}
    \toprule
    \textbf{SSH} & \textbf{FDA} & \textbf{IoU} & \textbf{mIoU} \\
    \midrule
    \midrule
    $\checkmark$ & & 47.18 & 16.53 \\
     & $\checkmark$ & 47.37 & 16.10 \\
    $\checkmark$ & $\checkmark$ & \textbf{47.91} & \textbf{17.01} \\
    \bottomrule
    \end{tabular}
    \label{tab:phe}
\end{table}

\subsection{Quantitative Results}
We conduct comparison experiments on two popular benchmarks, SemanticKITTI and SSCBench-KITTI-360, to evaluate the performance of our SPHERE against SOTA camera-based SSC methods.
\\
\textbf{SemanticKITTI:} Table~\ref{kitti} presents the comparison results on the SemanticKITTI validation set. It can be observed that our SPHERE approach obtains superior performance of \textbf{47.91\%} IoU for class-agnostic scene completion (SC) and \textbf{17.01\%} mIoU for semantic scene completion (SSC), respectively. Compared to the best comparison method SGN~\cite{mei2024camera} with ResNet50 backbone, our SPHERE achieves performance improvements of \textbf{1.70\%} IoU and \textbf{1.69\%} mIoU. Furthermore, compared with CGFormer~\cite{yu2024context}, which employs EfficientNetB7 as backbone and Swin-T~\cite{liu2021swin} as TPV feature backbone, our SPHERE still improves the SSC results by  \textbf{1.92\%} IoU and \textbf{0.14\%} mIoU. The comparison results showcase the effectiveness of our SPHERE in generating more realistic SSC predictions through joint exploitation of semantic and physical information.
\\
\textbf{SSCBench-KITTI-360:} Table~\ref{kitti360} demonstrates the comparison results on the SSCBench-KITTI-360 test set, where our SPHERE approach outperforms all comparison methods with the performance of \textbf{48.59\%} SC IoU and \textbf{20.56\%} SSC mIoU, respectively. It is worth noting that Gaussian-based methods~\cite{huang2024gaussianformer, huang2024probabilistic} utilize a total of 38400 Gaussians to represent the scene, resulting in unsatisfying 
performance due to spatial redundancy. On the other hand, our SPHERE selects focal anchors for Gaussian initialization, achieving superior performance with only 1024 Gaussians around focal regions.

\begin{table}[t]
    \centering
    \caption{Efficiency evaluation against voxel-based method SGN~\cite{mei2024camera} and Gaussian-based method GaussianFormer-2~\cite{huang2024probabilistic}.}
    \begin{tabular}{c|ccc}
    \toprule
    \textbf{Methods} & \textbf{Params (M)} & \textbf{Memory (G)} & \textbf{GFLOPs} \\
    \midrule
    \midrule
    GaussianFormer-2 & 71.57 & 39.21 & 1814.99 \\
    SGN & 28.16 & 15.83 & 725.05 \\
    \textbf{SPHERE~(Ours) }& 28.79 & 16.13 & 841.59 \\
    \bottomrule
    \end{tabular}
    \label{tab:efficiency}
\end{table}

\begin{figure}[t]
\centering
\includegraphics[width=1.0\columnwidth]{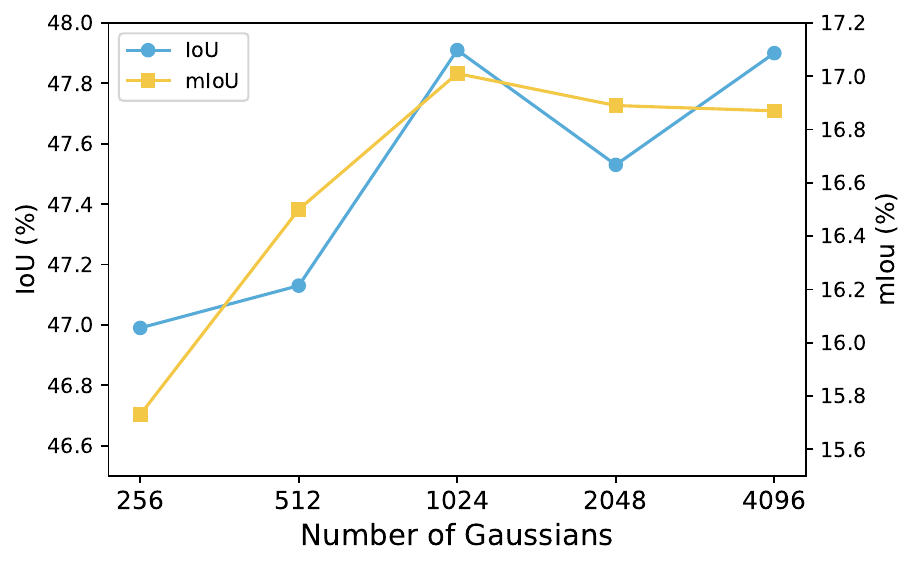} 
\caption{Effect of the number of Gaussians on the SSC performance for the SemanticKITTI validation set.}
\label{fig:gs}
\end{figure}

\begin{figure*}[t]
\centering
\includegraphics[width=\textwidth]{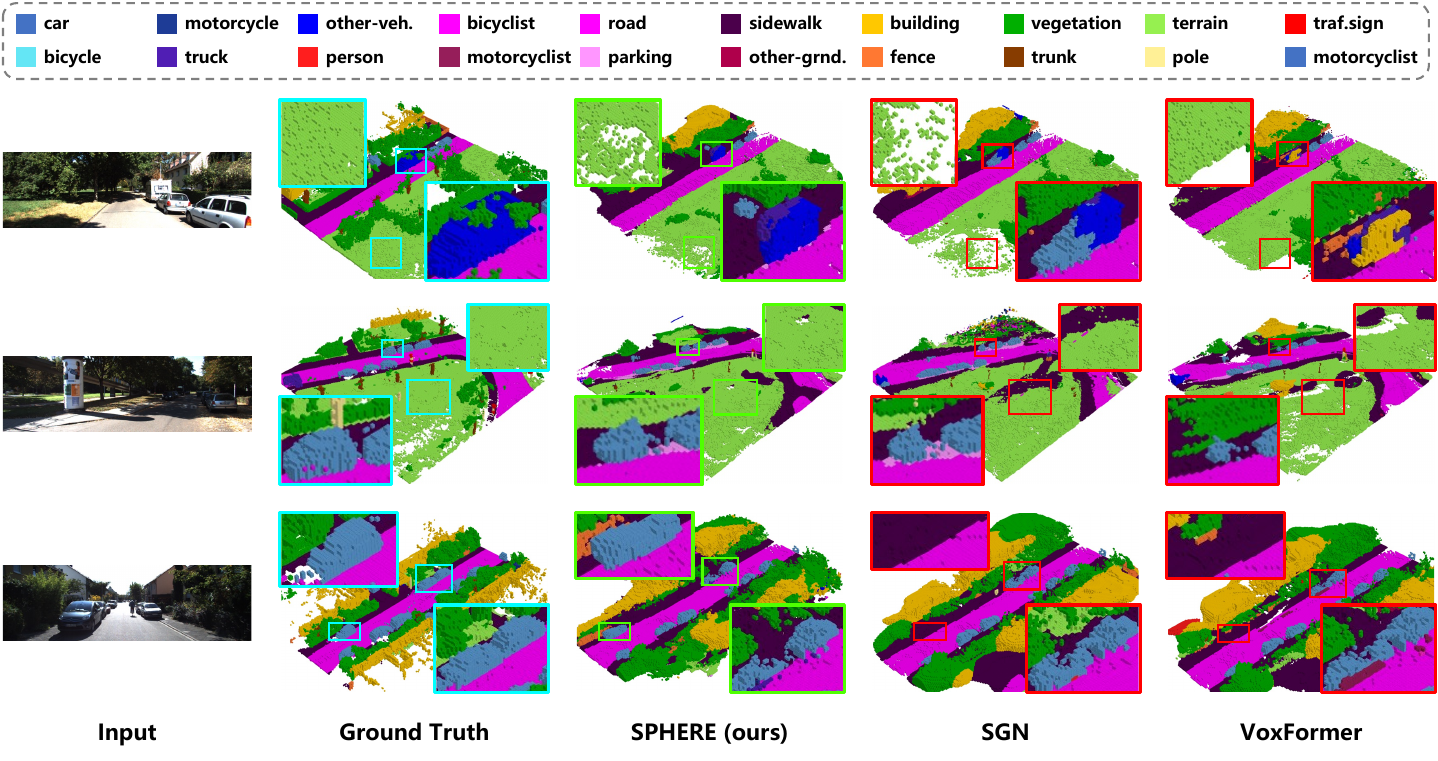}
\caption{Qualitative visualization results on the SemanticKITTI validation set. Cyan boxes outline the occupancy ground truth. Red boxes indicate false occupancy predictions of the best comparison method SGN and baseline method VoxFormer, while green boxes indicate improved scene completion results generated by our SPHERE approach. Better viewed when zoomed in.}
\label{fig:vis}
\end{figure*}

\subsection{Ablation Studies}
To further investigate the effectiveness of our SPHERE approach and different components, we conduct ablation experiments on the SemanticKITTI validation set as follows:
\\
\textbf{Ablation on Network Components.} As shown in Table~\ref{tab:ablation}, we adopt VoxFormer~\cite{li2023voxformer} as the baseline method, which achieves $44.15\%$ IoU and $13.5\%$ mIoU. By adding SGI to the baseline, the model performances are improved by $2.08\%$ IoU and $2.02\%$ mIoU respectively, which showcases the effectiveness of leveraging Gaussian representations for improved geometric details. Furthermore, by incorporating PHE, SPHERE is able to model physical-aware contextual details with semantic spherical harmonics and promotes semantic-geometry consistency via focal distribution alignment. Therefore, the SSC performance is further enhanced with improvements of $3.90\%$ IoU and $3.50\%$ mIoU over the baseline method.
\\
\textbf{Ablation on Physical-aware Harmonic Enhancement.} Tabel~\ref{tab:phe} presents the ablation experiments on SSH and FDA. It can be observed that incorporating SSH achieves higher performance improvements on the mIoU metric, since SSH focuses on capturing physical-aware contextual structures for more realistic details. On the other hand, integrating FDA better improves the IoU metric, since the aligned distribution promotes semantic-geometry consistency for more accurate predictions.
\\
\textbf{Efficiency Evaluation.} Table~\ref{tab:efficiency} compares the training efficiency of our SPHERE against voxel-based and Gaussian-based methods. It can be observed that our SPHERE achieves a balance between memory for voxel representations and computation for Gaussian representations, generating accurate semantics and realistic geometry efficiently and effectively.
\\
\textbf{Effect of Gaussian numbers.} Figure~\ref{fig:gs} illustrates the effect of the number of Gaussians $K$ in SGI. The best performance is obtained with $K=1024$, much fewer than that in Gaussian-based methods~\cite{huang2024gaussianformer, huang2024probabilistic}. This validates SPHERE alleviating the spatial redundancy issue with focal anchors for Gaussian initializations.

\subsection{Qualitative Results}
Figure~\ref{fig:vis} illustrates the visualization results on the SemanticKITTI validation set generated by VoxFormer~\cite{li2023voxformer}, SGN~\cite{mei2024camera}, and our proposed SPHERE. We highlight the ground truth with cyan boxes as a reference for comparison. The red boxes indicate false predictions of SGN and VoxFormer, while the green boxes highlight improved results from our SPHERE approach. It can be observed that SPHERE generates improved SSC predictions with more accurate semantic predictions and more realistic geometric structures, especially for classes like car, other-vehicle, vegetation, and road. This improvement is attributed to the physical-aware modeling ability of Gaussian representations, and the semantic-geometry consistency promoted by focal distribution alignment.

\section{Conclusion}
In this work, we focus on a key challenge in camera-based SSC: voxel-based methods struggle to learn physical regularities for realistic details, while neural reconstruction methods yield confusing semantics due to spatial redundancy in autonomous driving scenes. To address this issue, we propose the SPHERE framework, which integrates voxel and Gaussian representations to jointly model semantic and physical information. Specifically, SGI leverages dual-branch representations to identify focal anchors for efficient and effective Gaussian initialization. Then, PHE incorporates the semantic spherical harmonics to model improved physical-aware details and ensures semantic-geometry consistency through focal distribution alignment between voxel and Gaussian representations. Extensive experiments on SemanticKITTI and SSCBench-KITTI-360 demonstrate the effectiveness of our SPHERE approach in achieving accurate semantics and realistic geometry.
\\
\textbf{Limitations.} SPHERE leverages focal anchors with discriminative semantics for Gaussian initialization, taking advantage of both voxel and Gaussian representations for improved SSC performance. However, real-world scenarios involve complex conditions such as occlusion, motion blur, and low light, where low-quality semantic features damage model performance. In future work, we plan to incorporate explicit physical priors for improved performance with low-quality semantic features under complex input conditions.

\begin{acks}
This work was supported by the grants from Beijing Natural Science Foundation (L247006) and the National Natural Science Foundation of China (62525201, 62132001, 62432001). 
\end{acks}

\bibliographystyle{ACM-Reference-Format}
\bibliography{main}


\end{document}